\documentclass[10pt,twocolumn,letterpaper]{article}
\usepackage{iccv}
\usepackage{url}
\usepackage{algorithm}
\usepackage[noend]{algpseudocode}
\usepackage{times}
\usepackage{epsfig}
\usepackage{graphicx}
\usepackage{amsmath}
\usepackage{amssymb}
\usepackage{subfigure}
\usepackage{siunitx}
\usepackage{booktabs}
\usepackage{xspace}
\usepackage{times}
\usepackage{epsfig}
\usepackage{color}
\usepackage{derivative}
\usepackage{multirow}
\usepackage[accsupp]{axessibility} 
\usepackage[toc]{appendix}

\newcommand{\Point}{\matr{p}}
\newcommand{\matr}[1]{\mathbf{#1}}     
\newcommand{\Aff}{\matr{A}}
\newcommand{\trans}{\text{T}}

\newcommand{\tabincell}[2]{\begin{tabular}{@{}#1@{}}#2\end{tabular}}
\renewcommand{\paragraph}[1]{\vspace{0.5em}\noindent\textbf{#1}}

\usepackage[pagebackref=true,breaklinks=true,colorlinks,bookmarks=false]{hyperref}

\usepackage[accsupp]{axessibility}
\iccvfinalcopy 

\begin{document}
\ificcvfinal\pagestyle{empty}\fi
\title{Adaptive Reordering Sampler with Neurally Guided MAGSAC}

\author{Tong Wei$^1$, Ji{\v{r}}{\'\i} Matas$^1$, and Daniel Barath$^2$ \\
$^1$ Visual Recognition Group, FEE, Czech Technical University in Prague \\$^2$ Computer Vision and Geometry Group, ETH Zurich \\
{\tt\small \{weitong, matas\}@fel.cvut.cz, danielbela.barath@inf.ethz.ch}}

\maketitle
\ificcvfinal\thispagestyle{empty}\fi

\begin{abstract}
    We propose a new sampler for robust estimators that always selects the sample with the highest probability of consisting only of inliers. 
    After every unsuccessful iteration, the inlier probabilities are updated in a principled way via a Bayesian approach.
    The probabilities obtained by the deep network are used as prior (so-called neural guidance) inside the sampler. 
    Moreover, we introduce a new loss that exploits, in a geometrically justifiable manner, the orientation and scale that can be estimated for any type of feature, \eg, SIFT or SuperPoint, to estimate two-view geometry. 
    The new loss helps to learn higher-order information about the underlying scene geometry. 
    Benefiting from the new sampler and the proposed loss, we combine the neural guidance with the state-of-the-art MAGSAC++.  
    Adaptive Reordering Sampler with Neurally Guided MAGSAC (ARS-MAGSAC) is superior to the state-of-the-art in terms of accuracy and run-time on the PhotoTourism and KITTI datasets for essential and fundamental matrix estimation.
    The code and trained models are available at \url{https://github.com/weitong8591/ars\_magsac}.
\end{abstract}


\section{Introduction}
\label{sec:intro}
Robust estimation of two-view geometry is a fundamental problem in numerous computer vision applications, \eg, wide baseline matching~\cite{pritchett1998wide,matas2004robust,mishkin2015mods}, multi-model fitting~\cite{isack2012energy,pham2014interacting,barath2018multi}, initial pose recovery of Structure-from-Motion~\cite{schoenberger2016sfm,schoenberger2016mvs} and Simultaneous Localization and Mapping (SLAM)~\cite{mur2015orb} pipelines. 
RANSAC (RANdom SAmple Consensus)~\cite{RANSAC}, and its recent variants~\cite{raguram2013usac,barath2019magsac,barath2020magsac++,ivashechkin2021vsac,barath2022learning} have been widely applied in practice due to efficiency, simplicity, and accuracy that makes them appealing in real-world scenarios.   
In brief, RANSAC works in iterations by selecting a subset of data points, estimating the model (\eg, relative pose of a camera pair), and measuring its quality as the number of points consistent with it (\ie, its inliers). 

Since the publication of RANSAC, several modifications have been proposed to improve its accuracy and speed, focusing on specific components of the original algorithm. 
One common way to increase the accuracy is considering realistic noise distributions in model scoring, rather than relying on inlier counting that essentially assumes uniform noise. 
MLESAC~\cite{torr2000mlesac} uses a maximum likelihood procedure to reason about the quality of each minimal sample model. 
While MLESAC can achieve better accuracy than standard inlier counting, it can be computationally expensive compared to the original RANSAC.
MSAC~\cite{torr2002bayesian} proposes the use of truncated $L_2$ loss, which has been shown to be equivalent to the maximum likelihood-based approaches, but it still heavily relies on a manually set threshold. 
More recently, MAGSAC~\cite{barath2019magsac} and MAGSAC++~\cite{barath2020magsac++} have been proposed to alleviate the dependence on manual threshold selection by marginalizing over an acceptable range of noise scales.
As shown in the recent survey~\cite{ma2021image}, MAGSAC++ is currently the most accurate RANSAC variant as in \cite{ivashechkin2021vsac}. 

\begin{figure*}[t]
	\centering
     \vspace{-1.5mm}
    \includegraphics[width = 1.98\columnwidth]{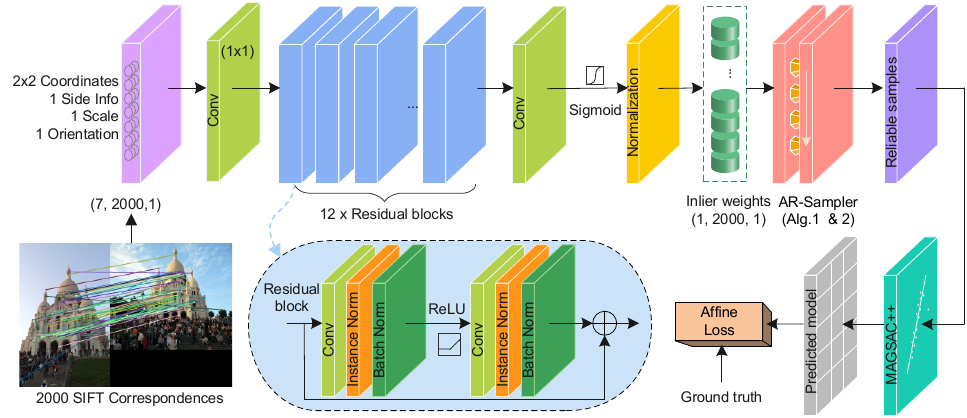}
    \caption{\textbf{ARS-MAGSAC.} 
    The point coordinates, SIFT orientations and scales and other information (\eg, SNN ratio~\cite{lowe1999object}) are fed into the network. 
    The predicted inlier probabilities used as priors are updated by the proposed AR-Sampler (see Sec.~\ref{sec:sampler}). 
    The estimated model, \eg relative pose, is used to calculate the loss that combines the proposed Affine Loss (see Sec.~\ref{subsec:affine}) and other ones.
   }
    \label{fig:flowchart}
    \vspace{-0.5em}
\end{figure*}

Improving the sampling procedure is another way to enhance the performance of RANSAC by selecting a good sample early and triggering the termination criterion.
Several samplers have been proposed, each with its own assumptions and limitations. 
The NAPSAC~\cite{torr2002napsac} sampler assumes that inliers are spatially coherent and, thus, it draws samples from a hyper-sphere centered at the first, randomly selected, location-defining point. 
The GroupSAC algorithm~\cite{ni2009groupsac} assumes that inliers are often ``similar'' and, thus, can be separated into groups. 
PROSAC~\cite{chum2005matching} exploits an a priori predicted inlier probability rank of each point and starts the sampling with the most promising ones.
Progressively, samples that are less likely to lead to the sought model are drawn.
More recently, \cite{brachmann2019neural} proposes a sampler to exploit prior knowledge of inlier probabilities, \eg, from a deep network. 
The sampler selects minimal samples according to the probability, assuming that it follows a categorical distribution over the discrete set of observations.

Recently, several algorithms have been proposed for robust relative pose estimation using neural networks. 
The first paper on the topic, Context Normalization Networks (PointCN)~\cite{cne2018} proposes the use of PointNet (MLP) with batch normalization as a context mechanism. 
Attentive Context Normalization Networks~\cite{acne2020} introduces a special architectural block for the task. 
The Deep Fundamental Matrix Estimation (DFE)~\cite{dfe2018} uses differentiable iteratively re-weighted least-squares with predicted weights.
The OANet algorithm~\cite{oanet2019} introduces differentiable pooling and unpooling blocks for correspondence filtering.
Neural Guided RANSAC (NG-RANSAC)~\cite{brachmann2019neural} uses a PointCN-like architecture with a different training objective, and the predicted correspondence scores are exploited inside RANSAC by using a guided sampling method that helps to find accurate models early.
More recently, CLNet~\cite{clnet2021} introduces several algorithmic and architectural improvements to remove gross outliers with iterative pruning.
These techniques provide alternatives for tentative correspondence pre-filtering and weighting. While these methods have shown promising results, they do not substitute standard robust estimation, as demonstrated in the RANSAC tutorial~\cite{cvpr2020ransactutorial}.

This paper makes three main contributions. 
\textit{First}, we propose a new sampler tailored for neurally guided robust estimators always selecting the sample with the highest probability of consisting only of inliers.
It exploits the inlier probabilities obtained by deep networks as prior knowledge and updates them in a principled way via a Bayesian approach.
\textit{Second}, we propose a new loss that incorporates the orientation and scale from off-the-shelf feature detectors, \eg, SIFT~\cite{lowe2004distinctive}, in a geometrically meaningful manner directly into the training. 
Additionally, we show that this extra information can be obtained for any features, \eg SuperPoint~\cite{detone2018superpoint}, allowing to seamlessly integrate the proposed loss with recent detectors.
The loss allows learning higher-order information about the underlying scene geometry and improves the robustness of the estimation in challenging environments.
\textit{Third}, as a technical contribution, we combine two state-of-the-art algorithms, \ie, MAGSAC++~\cite{barath2020magsac++} and NG-RANSAC~\cite{brachmann2019neural}, to improve the relative pose estimation accuracy on a wide range of scenes.

\section{Adaptive Re-ordering Sampler}
\label{sec:sampler}

In this section, we describe the proposed new sampler that always selects the sample with the highest probability of containing only inliers. 
This probability is updated adaptively according to the success or failure of the current minimal sample in robust estimation.
The new sampler will be called \textbf{AR-Sampler} in the remainder of the paper. 

Let us assume that we are given point correspondences $p_{i_1^t}$, $p_{i_2^t}$, $\dots$, $p_{i_n^t} \in \mathcal{P}$ with inlier probabilities $\mu_{i_1^t}$, $\mu_{i_2^t}$, $\dots$, $\mu_{i_n^t} \in [0, 1]$ such that $\mu_{i_1^t} \geq \mu_{i_2^t} \geq \dots \geq \mu_{i_n^t}$, where $i_1^t, \dots, i_n^t \in [1, n]$ are indices in the $t$th RANSAC iteration ensuring that the points are ordered by the inlier probabilities in a descending order.
The probability of sample $S = (p_{j_1}, p_{j_1}, \dots, p_{j_m}) \in \mathcal{P}^\times$ consisting only of inliers is calculated as $\mu_S = \prod_{k=1}^m \mu_{j_k}$, assuming independence, where $m$ is the sample size, \eg, $m = 5$ for essential matrix estimation. The independence assumption is incorrect for samples that include any subset of points that have previously been tested in another minimal sample, but it turns out to be a tractable and useful approximation. 
Consequently, the \textit{globally} optimal sampler maximizes the sample probability of $S^*_t = (p_{i_1^t}, p_{i_1^t}, \dots, p_{i_m^t})$ in the $t$th iteration to increase the probability of finding the sought model early.

\begin{algorithm}[ht]
\begin{algorithmic}[1]
	\Statex{\hspace{-1.0em}\textbf{Input:} $p_{1}, \dots, p_{n}$ -- points; $\mu_{1}, \dots, \mu_{n}$ -- probabilities }
	\Statex{\hspace{-1.0em}\phantom{\textbf{Input:}} $S$ -- minimal sample; $n_1, \dots, n_p$ -- usage numbers}
	\Statex{\hspace{-1.0em}\phantom{\textbf{Input:}} $(a_1, b_1), \dots, (a_n, b_n)$ -- initial distribution params}
    \Statex{\hspace{-1.0em}\textbf{Output:} $\mu_{1}', \dots, \mu_{n}'$ -- updated inlier probabilities}
   	\Statex{}
    \For{$i \in [1, n]$}
        \If{$p_i \in S$}\Comment{Decrease $\mu$ for all sampled points}
            \State{$a_i' \leftarrow a_i$; $b_i' \leftarrow b_i + n_i$}
            \State{$\mu_{i}' \leftarrow a_i' / (a_i' + b_i')$}
        \Else \Comment{Other points have the same $\mu$ as before}
            \State{$\mu_{i}' \leftarrow \mu_{i}$}
        \EndIf
    \EndFor
\end{algorithmic}
\caption{\bf Probability Update.}
\label{alg:sampler_update}
\end{algorithm}

\begin{algorithm}[ht]
\begin{algorithmic}[1]
	\Statex{\hspace{-1.0em}\textbf{Input:} $p_{1}, \dots, p_{n}$ -- points; $\mu_{1}, \dots, \mu_{n}$ -- inlier probs.}
	\Statex{\hspace{-1.0em}\phantom{\textbf{Input:}} $m$ -- sample size; $n_1, \dots, n_p$ -- usage numbers}
	\Statex{\hspace{-1.0em}\phantom{\textbf{Input:}} $(a_1, b_1), \dots, (a_n, b_n)$ -- initial distribution param.}
    \Statex{\hspace{-1.0em}\textbf{Output:} $S^*$ -- minimal sample}
   	\Statex{}
   	\State{$i_1, \dots, i_n \leftarrow \text{reorder}(\mu_{1}, \dots, \mu_{n})$ }\Comment{By the inlier prob.}
   	\State{$S^* \leftarrow \{ p_{i_j} \; | \; j \in [1, m] \}$}
   	\For{$p_{i_j} \in S^*$}
   	    \State{$n_{i_j} \leftarrow n_{i_j} + 1$}
   	    \State{$\mu_{i_j} \leftarrow \text{Update}(a_{i_j}, b_{i_j}, \mu_{i_j}, n_{i_j})$} \Comment{Algorithm~\ref{alg:sampler_update}}
   	\EndFor
\end{algorithmic}
\caption{\bf Adaptive Re-ordering Sampler.}
\label{alg:sampler}
\end{algorithm}
Every unsuccessful RANSAC iteration reduces the inlier probability of the points in the minimal sample. 
This stems from the fact that in the case of having an all-inlier sample that is good enough to find the sought model, RANSAC terminates.\footnote{Precisely, RANSAC either terminates immediately when it finds the sought model or the confidence exceeds the manually set threshold.  }
The sample that does not trigger the termination is not all-inlier (failure). Therefore, of all the possible inlier-outlier configurations in the sample, the ``all points are inliers'' is ruled out, and, consequently, the inlier probabilities of the sample points decrease, typically very modestly. 
In order to model this in a principled way, we update the probabilities using the Bayesian approach after each RANSAC iteration. We note that the Bayesian approach ignores the dependencies between points that appeared in a sample. 
As prior knowledge, we can either consider the output of the deep network or even the point ordering that the SNN ratio~\cite{lowe1999object} implies. 
In each update, only the points from the current sample are considered, and, thus, the probability of other points remains unchanged in the $t+1$-th iteration as we did not gather additional information about them. 

The probability of point $\textbf{p}$ being inlier in the $t$th iteration follows the Bernoulli distribution.
Consequently, the number of times point $\textbf{p}$ is being selected in an outlier-contaminated sample when selected $n_p$ times follows the binomial distribution with parameters $\mu_p (n_p)$ and $n_p$. 
The usual conjugate prior for a binomial distribution is a beta distribution with prior hyper-parameters  $a(n_p)$ and $b(n_p)$, with expectation
$
    a(n_p) / (a(n_p)+b(n_p))
$, 
variance
\begin{equation*}
    v = \frac{a(n_p)b(n_p)}{(a(n_p)+b(n_p))^2 (a(n_p)+b(n_p)+1)},
\end{equation*}
and posterior hyper-parameters $a(n_p)$ and $b(n_p)$.
The posterior distribution parameters are 
$a(n_p + 1) = a(n_p)$, $b(n_p + 1) = b(n_p) + 1$.
Since $a(n_p)$ is constant, we will simply write $a$ in the rest of the paper.   
The best estimator for $\mu_p(n_p + 1)$ using a quadratic loss function is an expectation of the posterior distribution. Consequently, 
\begin{equation}
    \mu_p(n_p + 1) = \frac{a}{a + b(n_p + 1)}.
\end{equation}

For each point $\textbf{p}$, the initial parameters of the beta distribution $a$ and $b(1)$ are set using the predicted inlier ratio $\mu_p(1) = \mu_p^{1}$.
We assume that the inlier probability prediction provides the expectation of the prior beta distribution and with the same mean precision for all points. 
Thus, the variance $v$ of all these initial beta distributions is equal and can be learned in advance.
Given the learned variance, 
\begin{eqnarray}  
    \mu_p(1)=\frac{a }{a + b(1)}, \,
    v=\frac{a b(1)}{(a + b(1))^2(a + b(1) + 1)}
\end{eqnarray}
leads to
\begin{eqnarray*}
    a = \frac{(\mu_p(1))^2(1-\mu_p(1))}{v}-\mu_p(1), \,
    b(1) = a  \frac{1-\mu_p(1)}{\mu_p(1)}.   
\end{eqnarray*}
Parameters $a $ and $b(1)$ are calculated prior to the robust estimation.
The probability update and the sampler are shown, respectively, in Algorithms \ref{alg:sampler_update} and \ref{alg:sampler}.
Both methods contain only a few calculations and, thus, are very efficient, shown in Sec.~\ref{sec:sampler}.
This is expected from a sampler in a RANSAC-like estimator where it runs in every iteration, often thousands of times. 
Note that we found that the sampler works better with probabilities shuffled by adding a small random number $\epsilon$.
Setting $\epsilon$ so it is uniformly distributed in-between $\pm5e^{-4}$ works well in all our experiments.

\section{Scale and Orientation Loss}
\label{sec:new_loss}

We propose a new loss function considering that, in most of the two-view cases, we apply feature detectors that provide more information about the underlying scene geometry than simply the point coordinates.
For instance, ORB~\cite{rublee2011orb} features contain the orientation of the image patches centered on the detected points in the two images.
In addition to the feature orientation, the SIFT~\cite{lowe1999object} and SURF~\cite{bay2006surf} detectors return a uniform scaling parameter.
Even the full affine warping of the patch can be recovered when using affine-covariant feature detectors, \eg, Hessian-Affine~\cite{mikolajczyk2005comparison} or MODS~\cite{mishkin2015mods}.
Moreover, even the most recent features, \eg SuperPoint~\cite{detone2018superpoint}, can be equipped with orientation and scale by applying the Self-Scale-Ori~\cite{lee2022self} method. 

This additional information that such features provide has not yet been exploited in a geometrically meaningful manner to minimize the training loss.
Recent deep networks, \eg,~\cite{brachmann2019neural}, use SNN ratios as side information added to the feature vectors or for pre-filtering correspondences.
\begin{figure}[t]
	\centering
    \includegraphics[width = 0.48\columnwidth,trim={0 10cm 0 0},clip]{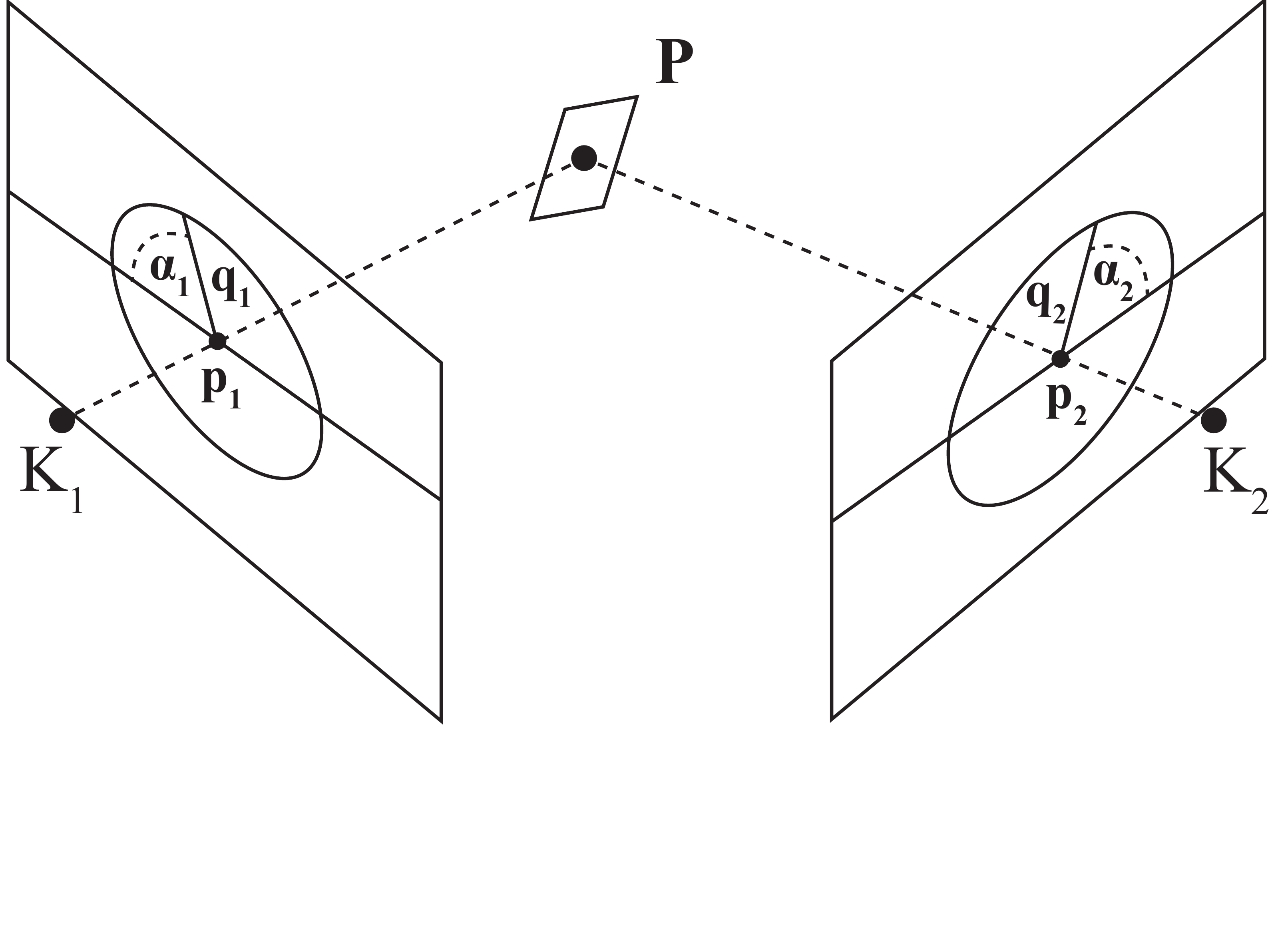}\hspace{2mm}
    \includegraphics[width = 0.48\columnwidth,trim={0 10cm 0 0},clip]{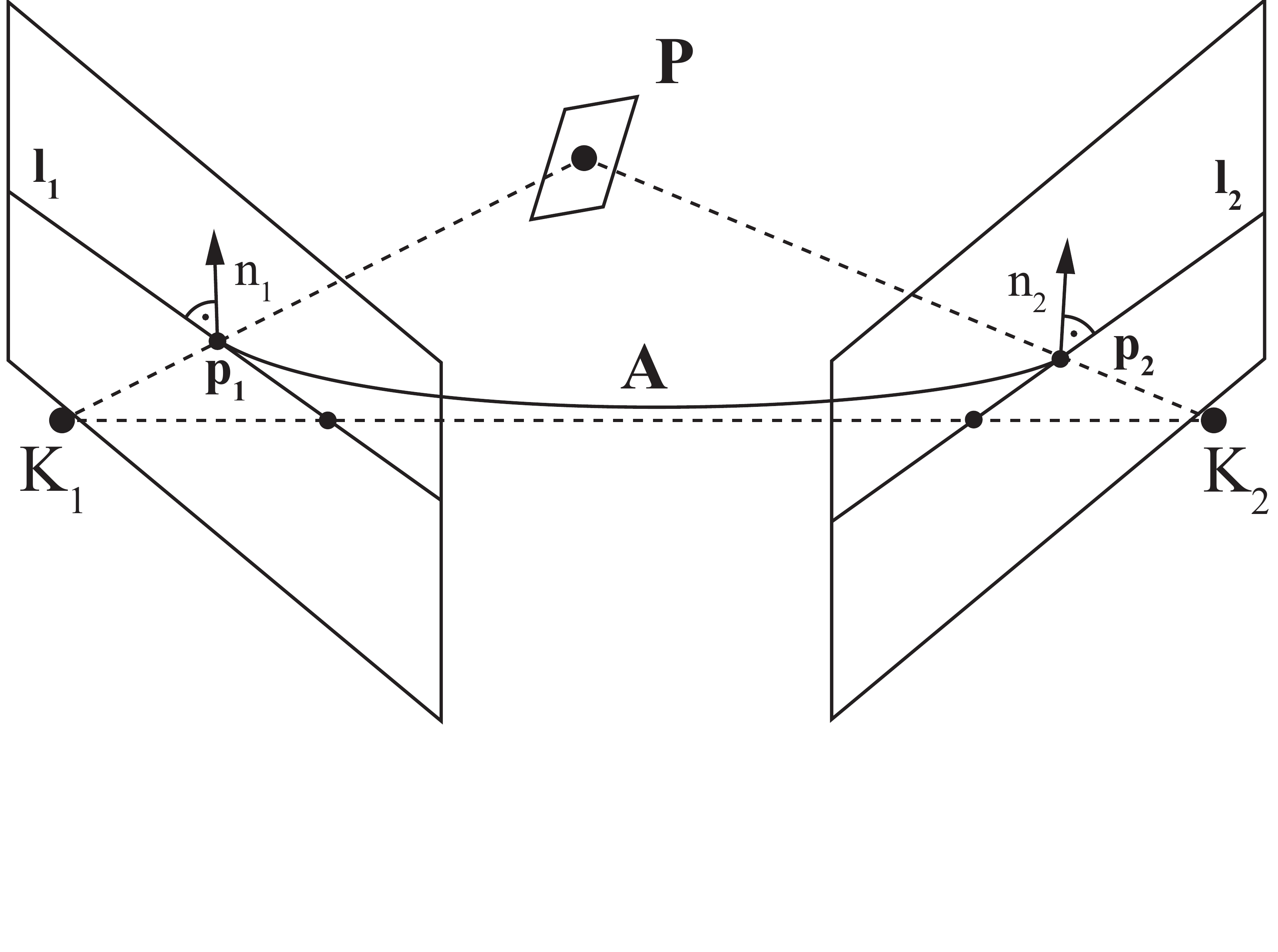}
    \caption{ \textbf{(Left)} Visualization of the orientation- and scale-covariant features. Point $\textbf{P}$ and the surrounding patch projected into cameras $\textbf{K}_1$ and $\textbf{K}_2$. 
	The rotation of the feature in the $i$th image is $\alpha_i \in [0, 2\pi)$ with size $q_i \in \mathbb{R}$, $i \in \{1,2\}$.
    \textbf{(Right)} The geometric interpretation of the relations of local affine transformations and the epipolar geometry (Eq.~\eqref{eq:affine_model}; proposed in \cite{barath2018efficient}). 
	The normal $\matr{n}_1$ of epipolar line $\matr{l}_1$ is mapped by affinity $\Aff \in \mathbb{R}^{2 \times 2}$ into the normal $\matr{n}_2$ of epipolar line $\matr{l}_2$. }
    \label{fig:geometric_interpr_sift}
    \vspace{-1.2em}
\end{figure}

\subsection{Affine Epipolar Error}
\label{sec:affine_loss}

In order to interpret fully or partially affine-covariant features, we adopt the definition from \cite{barath2018efficient} and the affine transformation model from \cite{barath2018five}. 
We consider an affine correspondence (AC) a triplet: $(\Point_1, \Point_2, \Aff)$, where $\Point_1 = [u_1, \quad v_1, \quad 1]^\trans$ and $\Point_2 = [u_2, \quad v_2, \quad 1]^\trans$ are a corresponding homogeneous point pair in the two images, and
\begin{eqnarray}
	\label{eq:affine_model}
    \begin{array}{lcl}
    	\Aff & = & \begin{bmatrix}
          a_{1}  & a_{2}  \\
          a_{3}  & a_{4} 
   		\end{bmatrix} = 
   		\begin{bmatrix}
          \cos(\alpha) & -\sin(\alpha) \\
          \sin(\alpha) & \cos(\alpha)
   		\end{bmatrix} 
   		\begin{bmatrix}
          q_{u} & w \\
          0 & q_{v}
   		\end{bmatrix}, 
    \end{array}
\end{eqnarray}
is a $2 \times 2$ linear transformation describing the relationship of the infinitesimal image patches centered on points $\Point_1$ and $\Point_2$, where $\alpha$ is rotation, $q_u$ and $q_v$ are the scales along the axes, and $w$ is the shear parameter. 
Formally, $\Aff$ is defined as the first-order Taylor-approximation of the $\text{3D} \to \text{2D}$ projection functions. 
For perspective cameras, $\Aff$ is the first-order approximation of the related $3 \times 3$ homography.

The relationship of affine correspondences and epipolar geometry is shown in \cite{raposo2016theory,barath2018efficient}, with \cite{barath2018efficient} providing a geometrically interpretable definition of the constraint as 
\begin{equation}
	\label{eq:normal_to_normal_1}
	\matr{A}^{-\trans} \matr{n}_1 = -\matr{n}_2
\end{equation} 
where $\matr{n}_1 = (\matr{F}^\trans \Point_2)_{[1:2]}$ and $\matr{n}_2 = (\matr{F} \Point_1)_{[1:2]}$ are the normals of the epipolar lines in the two images, and lower-index $\matr v_{[1:2]}$ selects the first two coordinates of a vector $\matr v$, as shown in the right plot of Fig.~\ref{fig:geometric_interpr_sift}.

While constraint \eqref{eq:normal_to_normal_1} is originally formulated as two linear equations in \cite{barath2018efficient} to simplify the estimation, it can be re-written to two geometrically meaningful constraints that we can use in the loss function. 
First, \eqref{eq:normal_to_normal_1} implies that $\matr{A}^{-\trans}$ rotates the normal in the first image to its corresponding pair in the second one as 
$
    (\matr{A}^{-\trans} \matr{n}_1) \times \matr{n}_2 = 0
$,
where the angle between $\matr{A}^{-\trans} \matr{n}_1$ and $\matr{n}_2$ can be used as an error for an estimated fundamental matrix $\matr{\widehat{F}}$ as follows:
\begin{equation}
    \label{eq:orientation_error}
    f(\Aff, \matr{\widehat{F}}, \Point_1, \Point_2) = \cos^{-1} \frac{\left(\matr{A}^{-\trans} (\matr{\widehat{F}}^\trans \Point_2)_{[1:2]} \right) (\matr{\widehat{F}} \Point_1)_{[1:2]}}{\left|\matr{A}^{-\trans}(\matr{\widehat{F}}^\trans \Point_2)_{[1:2]} \right| \left|(\matr{\widehat{F}} \Point_1)_{[1:2]} \right|}. 
\end{equation}
Second, \eqref{eq:normal_to_normal_1} implies that the scale change is 
\begin{equation}
    \sqrt{\det \Aff} = \frac{|\matr n_2|}{|\matr n_1|} = \frac{\left|(\matr{F} \Point_1)_{[1:2]} \right|}{\left|(\matr{F}^\trans \Point_2)_{[1:2]} \right|}
\end{equation}
providing another geometrically meaningful error as
\begin{equation}
    \label{eq:scale_error}
    g(\Aff, \matr{\widehat{F}}, \Point_1, \Point_2) = \sqrt{\det \Aff} - \frac{\left|(\matr{\widehat{F}} \Point_1)_{[1:2]} \right|}{\left|(\matr{\widehat{F}}^\trans \Point_2)_{[1:2]} \right|}. 
\end{equation}
Overall, these errors are used to measure the quality of the epipolar geometry given an affine correspondence. 

\subsection{Affine Loss Function}
\label{subsec:affine}
In practice, we are usually given partially affine-covariant features, \eg, with orientation and scale, that do not allow using \eqref{eq:orientation_error} and \eqref{eq:scale_error} directly. 
To define a justifiable loss, we first approximate the local affine frame $\widehat{\Aff}$ using the rotations $\alpha_1$, $\alpha_2$ and scales $q_1$, $q_2$ from the features via the affine transformation model in \eqref{eq:affine_model} assuming that shear $w = 0$, rotation $\alpha = \alpha_2 - \alpha_1$, and $q_u = q_v = q_2 / q_1$ is a uniform scaling along the axes similarly as in~\cite{barath2018five}, see the left plot of Fig.~\ref{fig:geometric_interpr_sift}.
It is important to note that directly using $\widehat{\Aff}$ to measure the error of the prediction is still not viable since $\widehat{\Aff}$ is only an approximation and, thus, \eqref{eq:orientation_error} and \eqref{eq:scale_error} are not zero even if the ground truth fundamental matrix is used. 
We, thus, define the orientation and scale losses respectively as
\begin{eqnarray*}
    L_\text{ori}({\dots}) & = & \left| f(\widehat{\Aff}, \matr{\widehat{F}}, \Point_1, \Point_2) - f(\widehat{\Aff}, \matr{F}, \Point_1, \Point_2) \right|, \\
    L_\text{scale}({\dots}) & = & \left| g(\widehat{\Aff}, \matr{\widehat{F}}, \Point_1, \Point_2) - g(\widehat{\Aff}, \matr{F}, \Point_1, \Point_2) \right|, 
\end{eqnarray*}
where $\matr{F}$ is the ground truth fundamental matrix used as a target for the network and $\matr{\widehat{F}}$ is the prediction. 
Measuring the error in this way allows ignoring the approximative nature of $\widehat{\Aff}$.
The final loss minimized is
\begin{small}
\begin{eqnarray*}
    L(\matr F, \matr{\widehat{F}}, \mathcal{P}) = \sum_{(\Point_1, \Point_2, \widehat{\Aff}) \in \mathcal{P}} w_\text{ori} L_\text{ori}(\matr{F}, \matr{\widehat{F}}, \widehat{\Aff}, \Point_1, \Point_2) + \\
    w_\text{scale} L_\text{scale}(\matr{F}, \matr{\widehat{F}}, \widehat{\Aff}, \Point_1, \Point_2) + \cdots
\end{eqnarray*}
\end{small}

\noindent
where $\mathcal{P} = \{ (\Point_1, \Point_2, \alpha_1, \alpha_2, q_1, q_2) \; | \; \Point_1, \Point_2 \in \mathbb{R}^2 \wedge \alpha_1, \alpha_2 \in [0, 2\pi] \wedge q_1, q_2 \in \mathbb{R}^+ \}$ is the set of correspondences, $w_\text{ori}$ and $w_\text{scale}$ are weighting parameters, and $\cdots$ represents other metrics, \eg, epipolar or pose error, or inlier ratio.
To propagate the gradient, the training objective $L(w)$ is defined as the minimization of the expected task loss, similarly as in \cite{brachmann2019neural}. Since integrating over all hypotheses to calculate the expectation is infeasible, the gradients for the categorical distribution over the discrete set of observations are approximated by drawing $K$ samples as
\begin{equation}\label{approx}
    \pdv{}{w}L(w) \approx \frac{1}{K} \sum^{K}_{k=1} [l(\hat{h})\pdv{}{w} \log p (H_k; w)],
\end{equation}
where $K$ is the number of samples used for gradient approximation. 
The task loss is $l(\hat{h})$ with $\hat{h}$ as the robust solver, and $p (H_k; w)$ is the learned distribution of the $k$th sample.

\section{Neurally Guided MAGSAC}
We combine NG-RANSAC~\cite{brachmann2019neural} and MAGSAC++~\cite{barath2020magsac++} with the proposed algprithms.
The pipeline is visualized in Fig.~\ref{fig:flowchart}. 
Even though we will describe it focusing on epipolar geometry estimation, ARS-MAGSAC is \textit{general}. 

MAGSAC++ formulates the robust estimation problem as an iteratively re-weighted least-squares (IRLS) approach.
Both the model quality calculation and inlier selection are done without making strict inlier-outlier decisions.
The model parameters $\theta_{i+1}$ in the $(i+1)$th step of the IRLS are calculated from the point-to-model residual function, $D(\theta_i,\textbf{p})$, where $\textbf{p}$ is a point from the input sets, as 
    $\theta_{i+1}= \text{arg min}_{\theta} \sum_{\textbf{p} \in \mathcal{P}} w( D(\theta_i,\textbf{p})) D^2(\theta,\textbf{p})$, 
where the weight of point $\textbf{p}$ is determined by marginalizing over the noise scale $\sigma$ as 
$w( D(\theta_i,\textbf{p}))=\int_{0}^{+\infty} \text{P}(\textbf{p} \; | \; \theta_i, \sigma)f(\sigma) {\mathrm d \sigma}$  
and $\theta_0 = \theta$, \ie, the initial model from the minimal sample. 

In order to improve MAGSAC++ using recent neural network-based techniques, we adopt the Neural Guided RANSAC (NG-RANSAC) architecture~\cite{brachmann2019neural}\footnote{While the architecture of NG-RANSAC is a simple MLP, it worked best in our experiments on predicting inlier probabilities for sampling.}.
The NG-RANSAC algorithm predicts the probability of each point correspondence being inlier and uses a weighted sampling approach to incorporate this information in the robust estimation procedure. 
Due to the neural network and the robust estimator being loosely connected in such a manner, we can replace RANSAC with MAGSAC++ with all its bells and whistles and retrain the network.
While training the weights with sparse correspondences end-to-end, the inlier masks and selected samples are used to update the gradients of the neurons and generate point probabilities as weights for the consequent epochs. We use additional side information as well, namely, the scale and orientation of each SIFT feature.

\section{Experimental Results}

We evaluate the accuracy and speed of ARS-MAGSAC and the impact of each individual improvement proposed in this paper, \eg, AR-Sampler and affine loss.
The compared methods are the OpenCV RANSAC~\cite{RANSAC} and LMEDS~\cite{rousseeuw1984least}, the implementations provided by the authors of GC-RANSAC \cite{barath2018graph}, MAGSAC \cite{barath2019magsac} and MAGSAC++~\cite{barath2020magsac++}, NG-RANSAC~\cite{brachmann2019neural}, OANet~\cite{oanet2019}, EAS~\cite{fan2021efficient}, and CLNet~\cite{clnet2021}.
We re-trained NG-RANSAC, OANet and CLNet using the same data as ARS-MAGSAC, explained in Sec.~\ref{sec:Etest} and~\ref{sec:Ftest}.
Also, we will show their results with the provided models trained on significantly more image pairs than what we use for training ARS-MAGSAC.
For fair comparison, we also run MAGSAC++ in the end of OANet and CLNet.
All the experiments were conducted on Ubuntu 20.04 with GTX 3090Ti, OpenCV 4.5/3.4, and PyTorch 1.7.1.

\vspace{1mm}\noindent{\textbf{Technical details.}}
We use RootSIFT~\cite{arandjelovic2012three} features to improve the robust estimation accuracy and help the deep network to learn accurate weights. 
RootSIFT is a strategy normalizing the SIFT~\cite{lowe1999object} descriptors, thus, helping the feature matcher to find good tentative correspondences.
When training, we provide the network with the feature scales and orientations as a learned side information. 
Also, we do SNN ratio~\cite{lowe1999object} filtering on the correspondences as a preliminary step. In the SNN test, the correspondences are discarded if the distance between the first and the second nearest neighbors is larger than a manually set threshold, which works well as we set to $0.8$ in all of our experiments.

\subsection{Essential Matrix Estimation}
\label{sec:Etest}

To test the essential matrix estimation of the proposed algorithm, we downloaded 13 scenes from the CVPR IMW 2020 PhotoTourism challenge \cite{snavely2006photo}.
These scenes were also used in the CVPR tutorial \textit{RANSAC in 2020} \cite{tutorialdata} to compare robust estimators.
The dataset contains tentative correspondences formed by mutual nearest neighbors matching RootSIFT descriptors, ground-truth intrinsic camera parameters and relative poses. 
We use scene St.\ Peter's Square, consisting of \num{4950} image pairs, for training ARS-MAGSAC, retraining NG-RANSAC, CLNet, and OANet, and tuning the hyper-parameters of other compared methods.
For testing, we use \num{1000} randomly chosen image pairs from each of the \num{12} scenes. 
Thus, the methods are tested on a total of \num{12000} image pairs.
For maximum reproducibility, we will provide these pairs together with the source code. 
\begin{table*}[t]
  \centering
  \vspace{-1.5mm}
\resizebox{1.92\columnwidth}{!}{\begin{tabular}{crcccccccccc}
    \toprule
\multicolumn{2}{r}{\multirow{2}{*}{Dataset / Method}} &\tabincell{c}{LMEDS} & \tabincell{c}{RSC} 
&  \tabincell{c}{ GC-RSC}& \tabincell{c}{MSC}& \tabincell{c}{MSC++} & \tabincell{c}{EAS} & \tabincell{c}{OANet~\cite{oanet2019}} &\tabincell{c}{CLNet~\cite{clnet2021}} &\tabincell{c}{NG-RSC} & \multirow{2}{*}{\tabincell{c}{ARS-MAGSAC}} \\
\multicolumn{2}{r}{}  &\tabincell{c}{\cite{rousseeuw2005robust}} & \tabincell{c}{\cite{RANSAC}} 
&  \tabincell{c}{\cite{barath2018graph}}& \tabincell{c}{ \cite{barath2019magsac}}& \tabincell{c}{
\cite{barath2020magsac++}} & \tabincell{c}{ \cite{fan2021efficient}} & \tabincell{c}{ + MSC++} &\tabincell{c}{ + MSC++} &\tabincell{c}{ \cite{brachmann2019neural}} & \\ 

\hline
\multicolumn{2}{r}{Avg. time (ms)~$\downarrow$} & $\textbf{26.7}$  & $88.1$  & $175.1$  & $239.4$ & $113.4$ & $325.8$ & $49.1$ & $57.5$ & $79.8$  & $33.9$  \\
\hline

\multirow{12}{*}{\rotatebox{90}{Relative Pose AUC@$10^\circ$~$\uparrow$}} & Buckingham P.  & $0.19$ & $0.20$ & $0.20$& $0.27$ & $0.26$  & $0.13 $& $0.19$ & $0.27$ &$0.28$&$\textbf{0.33} $ \\

& Brandenburg G. & $0.34$ & $0.42$ & $0.48$& $0.53$ & $0.54$ &$0.38$ & $0.49$ & $0.59$ & $0.55$& $\textbf{0.61}$\\ 

& Colosseum E. & $0.25$ & $0.25$ & $0.27$& $0.32$ & $0.31$ &$0.19$ & $0.29$& $\textbf{0.38}$ & $0.32$ & $0.36$\\ 

& Grand Place B. & $0.14$ & $0.14$ & $0.17$& $0.22$& $0.21 $& $0.10$&$0.19$ & $0.24$& $0.22$& $\textbf{0.32}$\\

& Notre Dame F. & $0.24$ & $0.27$ & $0.38$& $0.40$ & $0.41$ &$0.24$& $0.38$ & $\textbf{0.51}$ & $0.34$  & $0.49$\\

& Palace of W. & $0.19$ & $0.31$ & $0.36$& $0.37$ & $0.37$ & $0.25$ & $0.22$ & $0.33$ & $0.38$ & $\textbf{0.43}$\\

& Pantheon E. & $0.49$ & $0.41$ & $0.48$& $0.62$ & $0.62$ & $0.33$&$0.52$ &$0.65$& $0.62$ & $\textbf{0.72}$ \\

& Prague Old T. & $0.10$ & $0.11$ & $0.12$& $0.16$ & $0.16$ & $0.07$& $0.10$ & $\textbf{0.20}$ & $0.17$& $\textbf{0.20}$\\

& Sacre Coeur & $0.52$ & $0.64$ & $0.68$& $0.71$ & $0.71$ &$0.65$ & $0.58$ &$0.69$ & $0.63$ &$\textbf{0.75}$\\ 

& Taj Mahal & $0.36$ & $0.48$ & $0.52$& $0.52$ & $0.55$ &$0.47$ & $0.48$ &$0.62$ &$0.55$ & $\textbf{0.67}$\\

& Trevi Fountain & 0.28 & 0.29 & 0.30& 0.37& 0.35 &0.22& 0.33 &\textbf{0.48} & 0.38& 0.43\\

& Westminster A. & $0.46$ & $0.36$ & $0.49$& $0.51$ & $0.51$ & $0.33$ &$0.43$ & $0.54$&$0.49$ &$\textbf{0.70}$\\ 
\cline{2-2}

& All & $0.30$ & $0.32$ & $0.37$ & $0.42$ & $0.42$ & $0.28$& $0.35$ & $0.46$ & $0.41$& $\textbf{0.50}$ \\

    \bottomrule \\
  \end{tabular} }
           \vspace{-0.5em}
  \caption{
  Essential matrix estimation on the PhotoTourism dataset~\cite{snavely2006photo}. We report the AUC scores~\cite{cne2018} thresholded at $10^{\circ}$ (higher is better) calculated from the pose error, \ie, the maximum of the relative rotation and translation errors in degrees.
  The first row shows the average run-times (ms).
  The last one reports the scores averaged over all scenes. 
  For RANSAC, GC-RANSAC, MAGSAC and MAGSAC++, we use the threshold as in~\cite{barath2021marginalizing}. Also, we tuned the threshold for EAS manually. 
  We trained OANet, CLNet, NG-RANSAC, and ARS-MAGSAC on the same datasets.
  The results with the pre-trained models provided by the authors are in Tab.~\ref{tab:clnet_compare_st}. 
  }
\vspace{-1.2em}
\label{tab:main_result_e}
\end{table*} 

Before applying end-to-end training, we initialize our model by minimizing the Kullback–Leibler divergence~\cite{van2014renyi} of the prediction and the target distribution using a 1000-epoch-long initial training process. 
To our experiments, this procedure improves the convergence speed of the end-to-end training.
For the main experiments, ARS-MAGSAC is trained on RootSIFT correspondences for 10 epochs, with inlier-outlier threshold upper bound set to $0.75$ pixel, 
Adam optimizer~\cite{kingma2014adam}, batch size of $32$, and $10^{-5}$ learning rate.

In each iteration of the training process, the pre-filtered correspondences are re-ordered according to the predicted weights, and MAGSAC++ with AR-Sampler is applied to estimate the \textbf{E} matrix. 
The pose error is calculated as the maximum of the rotation
    $\epsilon_{\widehat{\textbf{R}}} = (180 / \pi) \cos^{-1} ( ( \text{tr} ( \widehat{\textbf{R}} \textbf{R}^\text{T} ) - 1 ) / 2)$,
and the translation errors
    $\epsilon_{\hat{\textbf{t}}} = (180 / \pi) \cos^{-1} \frac{\textbf t^\text{T} \hat{\textbf t}}{|\textbf t| |\hat{\textbf t}|}$,
in degrees, where $\widehat{\textbf{R}} \in \text{SO}(3)$ is the 3D rotation and $\hat{\textbf{t}} \in \mathbb{R}^3$ is the translation, both decomposed from the estimated $\widehat{\matr{E}}$. 
Note that we use the angular translation error since the length of $\hat{\textbf{t}}$ can not be recovered from two views~\cite{hartley2003multiple}. 
Also, note that the scale and orientations of the features have to be normalized together with the point correspondences by the intrinsic camera matrices $\textbf{K}_1$ and $\textbf{K}_2$ as proposed in~\cite{barath2019homography}.
The rotation remains unchanged.
The scale is normalized by $f_2 / f_1$, where $f_i$ is the focal length of the $i$th camera.

We adopted the neural network from \cite{brachmann2019neural} and \cite{cne2018}, a commonly used neural network for geometry data, which comprises 12 residual blocks that connect information from different layers and several multi-layer perceptions (MLPs). 
Each block is constructed by two linear layers, a batch normalization layer, and a ReLU activation function \cite{he2015delving}. Besides, the global context is included by adding the instance normalization \cite{ulyanov2016instance} layer into each block. The inlier probabilities of the matches are mapped by a Sigmoid function. Finally, MAGSAC++ with the proposed AR-Sampler estimates \textbf{E} with its iteration number fixed to $1000$.
\begin{figure*}[t]
    \centering
        \includegraphics[width=0.62\columnwidth]{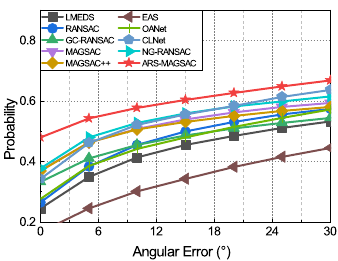}
        \includegraphics[width=0.62\columnwidth]{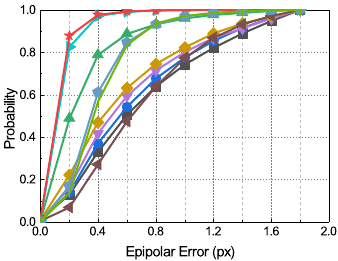}
       \includegraphics[width=0.62\columnwidth]{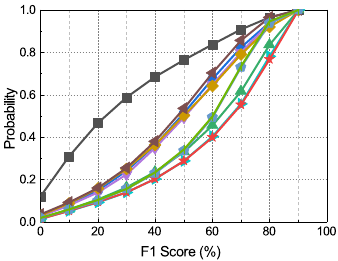}
    \caption{ The cumulative distribution functions (CDF) of the angular errors (left) for $\matr{E}$ estimation; epipolar errors (middle; in pixels) and F1 score (right; in percentages) for $\matr{F}$ estimation. 
    Essential matrix estimation was tested on \num{12000} image pairs from the PhotoTourism~\cite{snavely2006photo} dataset. 
    Fundamental matrix estimation is tested on \num{9690} pairs from the KITTI~\cite{geiger2012we} dataset. 
    We use the thresholds as in \cite{barath2021marginalizing} for the traditional algorithms. 
    We trained OANet, CLNet, NG-RANSAC, and ARS-MAGSAC on the same datasets.
    In the left two plots, being close to the top-left corner indicates accurate results. 
    In the right one (F1 score), the bottom-right corner is preferable. 
    }
    \vspace{-1em}
    \label{fig:cdf}
\end{figure*}

The training objective is defined as the minimization of the expected task loss~\cite{brachmann2019neural}. We approximate the gradients for the categorical distribution over the discrete set of observations, shown in Eq.~\ref{approx}.
The number of times each correspondence was selected in a minimal sample is back-propagated and used to update the weights and contribute to distribution learning for the next iteration.

\paragraph{PhotoTourism~\cite{snavely2006photo}, RootSIFT~\cite{arandjelovic2012three}.}
To measure the accuracy of the estimated essential matrices, we decompose them to rotation and translation and calculate the pose error.
Finally, we calculate the AUC scores at $5^\circ$, $10^\circ$ and $20^\circ$ from the pose errors as the area under the recall curves~\cite{brachmann2019neural}. 
The AUC@$10^\circ$ scores on each scene from the PhotoTourism dataset~\cite{snavely2006photo} are reported in Tab. \ref{tab:main_result_e}. Also, we show the run-time (in milliseconds) and the AUC scores averaged over all scenes.
The proposed ARS-MAGSAC is superior to the state-of-the-art on \textit{all} 
Scenes, \ie EAS, MAGSAC++, CLNet and NG-RANSAC, by a large margin both in terms of run-time and accuracy. 
Its average score is higher by $4$ AUC points than that of the second most accurate method (\ie, CLNet with MAGSAC++). 
The only faster method is LMEDS that has the second lowest accuracy.
The left plot of Fig.~\ref{fig:cdf} shows the cumulative distribution functions (CDF) of the pose errors of the estimated \textbf{E} matrices on all tested scenes. 
Being accurate is indicated by a curve close to the top-left corner.  
ARS-MAGSAC is significantly more accurate than the other compared methods. 
\begin{table}[ht]
\centering
\vspace{-1.5em}
\setlength{\tabcolsep}{3.5pt}
\resizebox{1.0\columnwidth}{!}{\begin{tabular}{lcccccccc}
\toprule 
Threshold & OANet \cite{oanet2019}& CLNet \cite{clnet2021} & NG-RANSAC \cite{brachmann2019neural} & ARS-MAGSAC\\ \midrule
@$5^\circ$ & $0.38$ & $0.41$ & $0.38$ & $\textbf{0.47}$ \\
@$10^\circ$ & $0.44$ & $0.48$ & $0.43$ &$\textbf{0.50}$ \\
@$20^\circ$  & $0.51 $&$ \textbf{0.56}$ & $0.49$ &$0.54$\\
\bottomrule \\
\end{tabular}}
         \vspace{-1em}
\caption{AUC scores~\cite{cne2018} of essential matrix estimation using the pre-trained models provided by the authors of NG-RANSAC, OANet and CLNet.
CLNet and OANet were trained on \num{541184} image pairs from the YFCC~\cite{thomee2016yfcc100m} dataset. 
NG-RANSAC was trained on \num{10000} pairs from scene St.\ Peter's Square of PhotoTourism dataset~\cite{snavely2006photo}.
ARS-MAGSAC was trained on \num{4950} pairs.
}
\label{tab:clnet_compare_st}
\vspace{-1.2em}
\end{table}

Furthermore, Tab.~\ref{tab:clnet_compare_st} shows the  performance comparison of ARS-MAGSAC with OANet, CLNet and NG-RANSAC when using the pre-trained models that their authors provide.
CLNet and OANet was trained on \num{541184} image pairs from the YFCC \cite{thomee2016yfcc100m} dataset. 
NG-RANSAC was trained on a total of \num{10000} pairs from the same scene as what we use for ARS-MAGSAC.
As a reminder, ARS-MAGSAC was trained on \num{4950} image pairs in total.
Even in this unfair comparison, the proposed method leads to the most accurate results in the AUC@$5^\circ$ and AUC@$10^\circ$ cases by a large margin -- in the AUC@$5^\circ$ case, it is better than CLNet (\ie, the second best) by \num{6} AUC points. 
This clearly shows that ARS-MAGSAC generalizes better than the state-of-the-art learning-based robust estimation approaches and is able to learn the underlying scene geometry better.

\paragraph{PhotoTourism~\cite{snavely2006photo}, SuperPoint~\cite{detone2018superpoint}.}
To demonstrate that any features can be made rotation and scale covariant by a post-processing step, we extracted SuperPoint features from the PhotoTourism dataset and used mutual NN matching to get correspondences. 
We estimated the feature orientations and scales by the recent Self-Scale-Ori method~\cite{woo2021global}.
We then trained two models, one with the proposed affine loss on the extracted orientations and scales, and one with the pose loss.
The \textbf{E} estimation results of these models and MAGSAC++ averaged over the 12 scenes of PhotoTourism are shown in Tab.~\ref{superpoint}.
ARS-MAGSAC trained with affine loss is superior to
training with only pose, showing that the proposed loss works with the state-of-the-art features as well. 

\begin{table}[t]
    \centering

   \resizebox{0.99\columnwidth}{!}{ \begin{tabular}{lcccccc}
    \toprule
    Method & Loss & AUC@$5^\circ$~$\uparrow$& AUC@$10^\circ$~$\uparrow$ & AUC@$20^\circ$~$\uparrow$ & Run-time (ms)~$\downarrow$\\
    \midrule
    MAGSAC++ &- & $0.378$ &$ 0.427$ & $0.480 $& $134.33$  \\  
       ARS-MAGSAC &Pose& $0.375$ &$0.433$ & $0.499$ & \phantom{1}$91.83$ \\ 
        ARS-MAGSAC &Affine&$\textbf{0.385}$ &  $\textbf{0.442}$ & $\textbf{0.509}$ & \phantom{1}$\textbf{85.83}$ \\
       \bottomrule \\
    \end{tabular}}
    \vspace{-0.5em}
    \caption{AUC scores and avg.\ run-times of MAGSAC++~\cite{barath2020magsac++} and the proposed ARS-MAGSAC on SuperPoint features~\cite{detone2018superpoint} on 12 test scenes from the PhotoTourism dataset.
Two versions of ARS-MAGSAC are shown, trained with the standard pose loss and the proposed affine loss on orientations and scales obtained by~\cite{woo2021global}.}
    \label{superpoint}
    \vspace{-0.8em}
\end{table}

\begin{figure}[t]
    \centering
        \includegraphics[width=0.495\columnwidth]{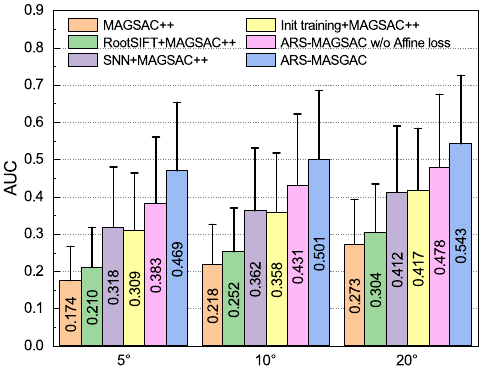}
        \includegraphics[width=0.495\columnwidth]{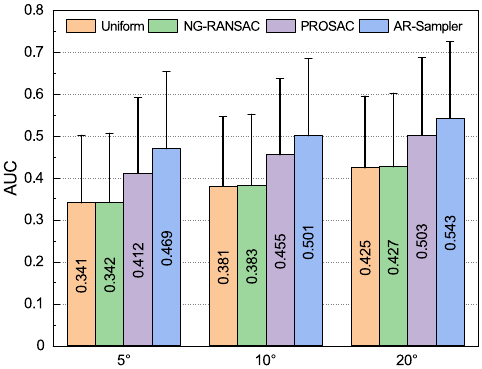}

    \caption{AUC scores at $5^\circ$, $10^\circ$ and $20^\circ$ of $\matr{E}$ estimation on the PhotoTourism dataset~\cite{snavely2006photo}.
    \textit{Left}:
    the impact of each component. 
    \textit{Right}: AUC scores testing with the Uniform~\cite{RANSAC}, NG-RANSAC sampler~\cite{brachmann2019neural}, PROSAC~\cite{chum2005matching} and the proposed AR-Sampler. 
    }
    \label{fig:technical}
    \vspace{-1.8em}
\end{figure}

\begin{table}[ht]
\centering
\resizebox{0.85\columnwidth}{!}{\begin{tabular}{ l | c c c }
        \hline   
           Threshold & Pose loss & Self-supervised loss & Proposed loss \\
        \hline    
          AUC@$5^\circ$~$\uparrow$& $0.38$  &  $0.41$ & $\textbf{0.47}$ \\
          AUC@$10^\circ$~$\uparrow$& $0.43$  &$ 0.45$ & $\textbf{0.50}$ \\
          AUC@$20^\circ$~$\uparrow$ & $0.48$  & $0.49$ & $\textbf{0.54}$ \\
        \hline  
\end{tabular}} \vspace{1mm}
\caption{AUC scores of essential matrix estimation using different loss functions for training ARS-MAGSAC on the RootSIFT features of the PhotoTourism benchmark.}
\label{tab:losses}
\vspace{-1.5em}
\end{table}

\subsection{Ablation Studies}
In the left plot of Fig.~\ref{fig:technical}, we show the accuracy gained from each component of the algorithm. 
We show the AUC scores and their std.\ at $5^\circ$, $10^\circ$ and  $20^\circ$ averaged over 12 scenes. 
The proposed affine loss plays an important role in improved accuracy. 
Also, it confirms that the widely used techniques, \eg, SNN filtering, RootSIFT, initial training, are important steps to achieve state-of-the-art results.

In the right plot of Fig.~\ref{fig:technical}, we show the results of different samplers used within ARS-MAGSAC.
The compared samplers are the uniform one from \cite{RANSAC}, the NG-RANSAC sampler~\cite{brachmann2019neural}, PROSAC~\cite{chum2005matching}, and the proposed AR-Sampler. 
It can be seen that the proposed AR-Sampler leads to the best accuracy.
Interestingly, PROSAC significantly outperforms NG-RANSAC, which is just marginally more accurate than the uniform sampler when used inside ARS-MAGSAC. 

The average AUC scores of essential matrix estimation when using the pose error as loss (Pose loss), outlier ratio as loss (Self-supervised loss), and the proposed one (when training ARS-MAGSAC) on the PhotoTourism dataset are shown in Tab.~\ref{tab:losses}. 
The proposed loss combining pose and affine losses leads to the most accurate results.

Additionally, Tab.~\ref{tab:networks_e} reports the AUC scores and the run-time of essential matrix estimation using our proposed method, or combined with other networks, \eg, the model proposed by CLNet~\cite{clnet2021}, replacing residual blocks with densely connected blocks among different layers, \etc. The proposed ARS-MAGSAC shows the best accuracies and marginally worse run-time than the best. Our architecture generalizes better than other networks, as simple as possible. The dense connections learn from the given correspondences as well as CLNet, but with low efficiency.
\begin{table}[ht]
    \centering
    \resizebox{1.0\columnwidth}{!}{\begin{tabular}{ l| c c c c}
            \hline   
              Threshold & \tabincell{c}{ARS-MAGSAC \\+ GNN layer~\cite{kipf2016semi}} & \tabincell{c}{ARS-MAGSAC \\+ CLNet model~\cite{clnet2021}}& \tabincell{c}{ARS-MAGSAC \\+ Dense blocks~\cite{Huang_2017_CVPR}} & ARS-MAGSAC \\
            \hline    
              AUC@5$^\circ$ & $0.37$  &  $0.41$ &$0.41$ & $\textbf{0.47}$ \\
              AUC@10$^\circ$ & $0.41$  & $0.45$ &$0.45$& $\textbf{0.50}$ \\
              AUC@20$^\circ$ & $0.46$  & $0.50$ &$0.50$ & $\textbf{0.54}$ \\ 
              Run-time (ms) & $\textbf{22.4}$ & $84.4$ & $127.8$&$31.4$\\
            \hline   
    \end{tabular}}\vspace{1.2mm}
    \caption{AUC scores of essential matrix estimation using ARS-MAGSAC with different networks on PhotoTourism.}
    \label{tab:networks_e}
    \vspace{-1em}
\end{table}

\subsection{Fundamental Matrix Estimation}
\label{sec:Ftest}
We test ARS-MAGSAC for \textbf{F} estimation on the KITTI benchmark \cite{geiger2012we}. 
As in~\cite{cne2018,brachmann2019neural}, Sequences ``00-05'' and ``06-10''  are regarded as the training and testing sets, respectively. 
The KITTI dataset consists of consecutive frames of high-resolution cameras rigidly mounted to a moving vehicle in a mid-size city, rural areas and highways \cite{geiger2012we}. 
The images are of size $1226 \times 370$. 
Correspondences are detected between subsequent images. 
In total, we use \num{14130} image pairs for training, and another \num{9060} for testing.

\begin{table}[ht]
\centering
\vspace{-1.2mm}
\setlength{\tabcolsep}{3.5pt}
\resizebox{0.99\columnwidth}{!}{\begin{tabular}{l c c c c c c c c c c}
\toprule 
Method & \tabincell{c}{LMEDS \\\cite{rousseeuw2005robust}} & \tabincell{c}{RSC \\\cite{RANSAC}} 
&  \tabincell{c}{ GC-RSC\\ \cite{barath2018graph}}& \tabincell{c}{MSC\\  \cite{barath2019magsac}}& \tabincell{c}{MSC++ \\
\cite{barath2020magsac++}} & \tabincell{c}{EAS \\ \cite{fan2021efficient}} & \tabincell{c}{OANet\\ \cite{oanet2019}} &\tabincell{c}{CLNet \\ \cite{clnet2021}}&\tabincell{c}{NG-RSC \\\cite{brachmann2019neural}} & \tabincell{c}{ARS-\\MSC} \\ \midrule
F1 score (\%)$\uparrow$ & $38.55$ & $56.83$ & $66.90$ & $57.80$ & $60.65$ & $55.16$ &$64.10$&$64.47$& $69.50$ &$\textbf{69.93}$  \\
AUC@$10^\circ$~$\uparrow$ & \phantom{1}$0.45$ & \phantom{1}$0.78$ & \phantom{1}$0.91$ & \phantom{1}$0.76$ & \phantom{1}$0.83$ & \phantom{1}$0.83$ &\phantom{1}$0.94$&\phantom{1}$0.95$& \phantom{1}$0.92$ & \phantom{1}$\textbf{0.97}$ \\
Error (px)~$\downarrow$ & \phantom{1}$3.15$ & \phantom{1}$0.98$ &\phantom{1}$0.42$ & \phantom{1}$0.84$ & \phantom{1}$0.75$ & \phantom{1}$0.88$ &\phantom{1}$0.57$&\phantom{1}$0.54$& \phantom{1}$0.41$ & \phantom{1}$\textbf{0.29}$ \\
Run-time (ms)~$\downarrow$ & \phantom{1}$20$ & \phantom{1}$32$ & \phantom{1}$56$ & \phantom{1}$233$ & \phantom{1}$413$ & \phantom{1}$310$ &\phantom{1}$17$&\phantom{1}$13$& \phantom{1}$18$ & \phantom{1}$\textbf{12}$ \\
\bottomrule \\
\end{tabular}}
\caption{The F1 score, AUC score thresholded at $10^{\circ}$, and median symmetric epipolar error (in pixels) of fundamental matrix estimation on \num{9690} images pairs from the KITTI benchmark~\cite{geiger2012we}.}
\label{tab:main_F_kitti}
\vspace{-1em}
\end{table}

Fundamental matrix estimation runs on the same architecture as what we described in Sec.~\ref{sec:Etest}. 
In this case, neither the point coordinates nor the orientations and scales are normalized. 
In contrast to \textbf{E} estimation, we do not apply initial training as it does not improve the accuracy here. 

\paragraph{KITTI~\cite{geiger2012we}, RootSIFT~\cite{arandjelovic2012three}.}
Tab. \ref{tab:main_F_kitti} reports the average run-time in milliseconds, the median symmetric epipolar error in pixels, the AUC and F1 scores of the estimated \textbf{F} matrices. 
ARS-MAGSAC leads to the highest accuracy in all metrics. 
Interestingly, while the F1 score is only marginally higher than that of NG-RANSAC, the AUC score is better by $5\%$. 
This implies that the F1 score is not in perfect agreement with the actual camera pose error captured in the AUC score. 
The run-time of ARS-MAGSAC is the lowest, being $33\%$ faster than NG-RANSAC. 
These timings exclude the prediction time which is at most $1-2$ milliseconds. 
Moreover, Fig.~\ref{fig:cdf} shows the CDFs of the epipolar errors (middle) and the F1 scores (right) on the \num{9060} image pairs. 
ARS-MAGSAC leads to the lowest errors and highest F1 scores. 

\paragraph{PhotoTourism~\cite{snavely2006photo}, RootSIFT~\cite{arandjelovic2012three}.}
We compare \textbf{F} matrix estimation on the RootSIFT features of the same scenes as we test for \textbf{E}. The F1 scores and the run-time are shown in Table~\ref{tab:photo_result_f}. 
The proposed ARS-MAGSAC leads to the best results on all but one scene, where it is the second best. Comparable rum-time is achieved among the lowest ones.
\begin{table}[ht]
\centering

\setlength{\tabcolsep}{3.5pt}
\resizebox{0.99\columnwidth}{!}{\begin{tabular}{l c c c c c c c c c c}
\toprule 
Method & \tabincell{c}{LMEDS \\\cite{rousseeuw2005robust}} & \tabincell{c}{RSC \\\cite{RANSAC}} 
&  \tabincell{c}{ GC-RSC\\ \cite{barath2018graph}}& \tabincell{c}{MSC\\  \cite{barath2019magsac}}& \tabincell{c}{MSC++ \\
\cite{barath2020magsac++}} & \tabincell{c}{EAS \\ \cite{fan2021efficient}} & \tabincell{c}{OANet\\ \cite{oanet2019}} &\tabincell{c}{CLNet \\ \cite{clnet2021}}&\tabincell{c}{NG-RSC \\\cite{brachmann2019neural}} & \tabincell{c}{ARS-\\MSC} \\ \midrule

AUC@$10^\circ$~$\uparrow$ & $35.00$ & $40.10$ & $43.41$ & $42.68$ & $42.46$ & $35.30$& $36.91$ & $40.67$ & $43.66$& $\textbf{47.76}$\\

Run-time (ms)~$\downarrow$ &$\textbf{21.00}$  & $30.67$  & $73.25$  & $281.30$ &$ 318.13$ & $325.83$ &$\textbf{21.00}$ &$34.83$& $25.85$  & $31.35$ \\
\bottomrule \\
\end{tabular}}
    \vspace{-0.4em}
\caption{Fundamental matrix estimation on the RootSIFT features of the PhotoTourism dataset~\cite{snavely2006photo}. Average run-times (ms) in the \textit{first} row, F1 scores on each scene and the average in the \textit{end}.
    For RANSAC, GC-RANSAC, MAGSAC and MAGSAC++, we use the threshold as in~\cite{barath2021marginalizing}. 
    We trained OANet, CLNet, NG-RANSAC on the same datasets as we use to train ARS-MAGSAC.}
\label{tab:photo_result_f}
\vspace{-1em}
\end{table}

\paragraph{PhotoTourism~\cite{snavely2006photo}, SuperPoint~\cite{detone2018superpoint}.}
In addition, we train and test ARS-MAGSAC for \textbf{F} matrix estimation on learning-based features on the PhotoTourism dataset. 
The coordinates are detected by SuperPoint~\cite{detone2018superpoint} and matched by the mutual nearest neighbor matcher.  
The feature orientations and scales are obtained by the state-of-the-art Self-Scale-Ori method~\cite{woo2021global}.
We trained on these features both with the pose error and our proposed affine loss that learns from the scales and orientations. 
Tab.~\ref{superpoint_f} demonstrates that the proposed method works accurately with SuperPoint features as well. The best performance is achieved with the proposed affine loss.

\begin{table}[ht]
\vspace{-0.5em}
\resizebox{0.99\columnwidth}{!}{
\begin{tabular}{l|ccccc}
    \hline
    Method & Loss  & F1 score (\%)~$\uparrow$ & med. epi. error (px)~$\downarrow$ & run-time (ms)~$\downarrow$\\
\hline    
       MAGSAC++~\cite{barath2020magsac++} &-& $26.65$ & $4.43$ & $180.50$ \\
        ARS-MAGSAC &Pose& $29.45$ &$4.30$ & \phantom{1}$23.92$\\
       ARS-MAGSAC &Affine& $\textbf{29.72}$ &  $\textbf{4.13}$ & \phantom{1}$\textbf{22.83}$ \\
     \hline
    \end{tabular}} \vspace{1.2mm}
     \caption{
  Fundamental matrix estimation on the SuperPoint features of the PhotoTourism dataset~\cite{snavely2006photo}. We show the trained models with two different losses, compared with MAGSAC++.}
  \vspace{-1.5em}
  \label{superpoint_f}
\end{table}
\paragraph{Comparison to \cite{bian2019evaluation}, RootSIFT.}
We compare the proposed ARS-MAGSAC with running LMEDS~\cite{rousseeuw2005robust} as a post-processing step on MAGSAC++~\cite{barath2020magsac++} or GC-RANSAC~\cite{barath2018graph} as proposed in \cite{bian2019evaluation} for \textbf{F} matrix  estimation.
We use the dataset from \cite{bian2019evaluation} and also PhotoTourism.
The F1 scores are reported in Table~\ref{tab:bian}.
ARS-MAGSAC is the most accurate on all but one scene (\ie, TUM~\cite{sturm2012benchmark}). 
On TUM, all methods lead to similar results and the differences are small. 

\begin{table}[ht]
\vspace{-0.5em}
    \hspace{2mm}\resizebox{0.95\columnwidth}{!}{\begin{tabular}{ l | c | c c | c c }
    \hline   
        Dataset & ARS-MSC & MSC++~\cite{barath2020magsac++} & + LMEDS~\cite{rousseeuw2005robust} & GC-RSC~\cite{barath2018graph} & + LMEDS \\
    \hline   
        CPC~\cite{wilson2014robust} & \textbf{27.40} & 25.18 & 25.32 & 24.56 &25.19\\
        KITTI~\cite{geiger2012we} & \textbf{69.93} & 60.65 & 69.45 & 66.90 & 69.50 \\
        T\&T~\cite{knapitsch2017tanks}&\textbf{14.04} & 13.58& 13.61& 12.58 & 12.82\\
        TUM~\cite{sturm2012benchmark} & \phantom{1}9.17 & \phantom{1}9.20 & \phantom{1}9.20 & \textbf{\phantom{1}9.24} & \phantom{1}9.21 \\
    \hline   
        PhotoTourism\ & \textbf{47.76} & 42.46 & 42.29 & 43.41 & 42.66\\
    \hline   
\end{tabular}}\vspace{1.2mm}
\caption{F1 scores (in percentages) of \textbf{F} estimation on the dataset from \cite{bian2019evaluation} using the proposed ARS-MAGSAC, and MAGSAC++ or GC-RANSAC followed by LMEDS as proposed in \cite{bian2019evaluation}. }
         \vspace{-1.5em}
\label{tab:bian}
\end{table}
\begin{figure}[ht]
\vspace{-0.5em}
\centering
    \includegraphics[width = 0.48\columnwidth,trim={0 0 0 0},clip]{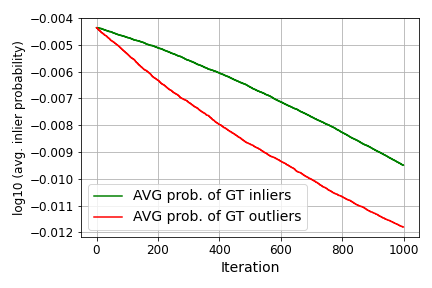}
    \includegraphics[width = 0.48\columnwidth,trim={0 0 0 0},clip]{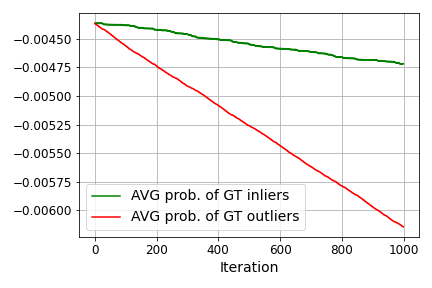}
    \caption{Avg.\ inlier probabilities of the GT inliers ({green}) and outliers ({red}) over iterations as updated by the proposed AR-Sampler. (\textit{Left}) \textbf{E} matrix estimation. (\textit{Right}) Absolute pose estimation. }
    \label{fig:arsampler}
    \vspace{-1em}
\end{figure}
\begin{figure*}[ht]
    \centering
    \vspace{0.5em}
        \includegraphics[width=0.62\columnwidth]{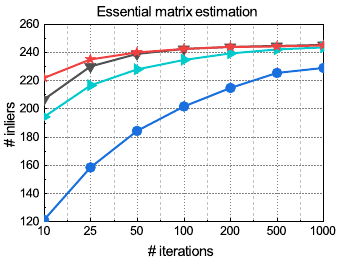}
        \includegraphics[width=0.62\columnwidth]{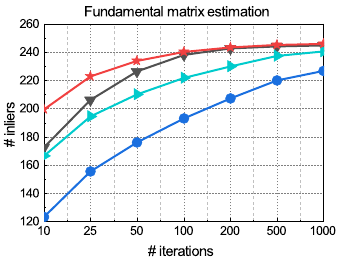}
        \includegraphics[width=0.62\columnwidth]{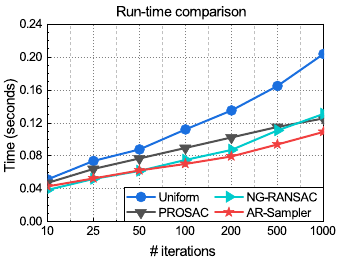}
    \caption{
    \textit{Left}, \textit{middle}: 
     the number of inliers (vertical axis) of the best model found within a given number of iterations (horizontal axis)  by MAGSAC++ ~\cite{barath2020magsac++} when combined with the Uniform~\cite{RANSAC} (blue), NG-RANSAC~\cite{brachmann2019neural} (green), PROSAC~\cite{chum2005matching} (black) and the proposed AR-Sampler (red). Average over \num{4950} image pairs from scene Sacre Coeur.
    \textit{Right}: the run-time, in seconds, versus the iteration number.
    The SNN ratio~\cite{lowe1999object} provides the inlier probabilities here, without using any learning algorithms.  
    }
    \label{fig:sampler_comparison}
\end{figure*}

\subsection{Sampler Comparison}
\label{sec:sampler}
\paragraph{Decreasing Speed Comparison.}
To give a more nuanced understanding of the proposed sampler, we ran essential matrix estimation on a randomly selected image pair and recorded the updated inlier probabilities throughout the iterations. 
These probabilities are then averaged independently for the ground-truth (GT) inliers (green curve) and outliers (red) and plotted in the \textit{left} plot of Fig.~\ref{fig:arsampler}. 
Additionally, to test the proposed AR-Sampler on a completely different problem, we ran the P1AC~\cite{ventura2022p1ac} single-point solver, estimating the absolute pose of a single query image (\textit{right} plot).
In both cases, the probabilities of the outliers w.r.t.\ the ground truth reduce faster than that of the inliers, demonstrating that the proposed sampler works as intended.

\paragraph{Sampler Comparison with SNN Ratio.}
We test the proposed sampler on the \num{4950} image pairs from scene Sacre Coeur when using the second nearest neighbor (SNN) ratio to order the points according to the inlier probabilities. 
To our experiments, considering the SNN ratio directly as prior inlier probability does not lead to an improvement compared to PROSAC. 
However, exploiting the point ranks implied by the SNN ratio works well.
Assume that we are given $n$ points $\textbf{p}_{i_1}, \dots, \textbf{p}_{i_n}$ ordered by their SNN ratios $s_{i_1}, \dots, s_{i_n}$. Thus, $s_{i_1} \leq s_{i_2} \leq \dots \leq  s_{i_n}$.
We calculate the prior probability of the $i_j$th point as $\mu_{i_j} = 1 - (j - 1) / (n - 1)$, $j \in [1, n]$.
Consequently, the first point, has 11 as prior probability when ordered by the SNN ratio.
Conversely, the last one is assigned zero.

The inlier numbers and run-times of the original MAGSAC++ when used together with the uniform~\cite{RANSAC}, PROSAC~\cite{chum2005matching}, NG-RANSAC samplers~\cite{brachmann2019neural}, and the proposed one are shown in Fig.~\ref{fig:sampler_comparison}.
The horizontal axis is the max.\ iteration number which is a strict upper bound on the iteration number that is controlled by the RANSAC confidence parameter.
The curve of the proposed sampler starts from a higher inlier number, both for \textbf{E} and \textbf{F} estimation, than that of the others, \ie, it leads to finding good samples earlier than the other methods.
As expected all methods converge to similar results after many iterations. Due to being extremely efficient, AR-Sampler leads to the fastest robust estimation, as shown in the right plot of Fig.~\ref{fig:sampler_comparison}.

\paragraph{Run-time of the Re-ordering Procedure.}
In the proposed adaptive re-ordering sampler, we use the priority queue implemented in the standard C++ library based on a heap structure to efficiently update the probabilities. The average run-time of the update is 32 microseconds in case of \textbf{E} matrix estimation ($m = 5$). For comparison, the PROSAC~\cite{chum2005matching} update costs 97 microseconds on average.

\section{Conclusion}

We propose ARS-MAGSAC, a novel algorithm for robust relative pose estimation that achieves state-of-the-art accuracy while maintaining comparable or better processing times than its less accurate alternatives. 
ARS-MAGSAC runs in real-time on most of the tested problems, making it highly practical for computer vision applications.
Additionally, we introduce a new loss that leverages additional geometric information, such as feature orientation and scale, improving the robustness in challenging environments. 
For many features, this extra information is available \textit{for free}.
For other ones, it can be easily extracted by, \eg, the Self-Scale-Ori method~\cite{lee2022self}. 
Furthermore, our proposed AR-Sampler outperforms traditional samplers, both when using predicted weights or SNN ratios as inlier probabilities. 
We also demonstrate that ARS-MAGSAC generalizes better than state-of-the-art learning-based approaches.

\paragraph{Acknowledgements.}
This research was supported by Research Center for Informatics (project CZ.02.1.01/0.0/0.0/16\_019/0000765 funded by OP VVV), by the Grant Agency of the Czech Technical University in Prague, grant No. SGS23/173/OHK3/3T/13, and by the ETH Postdoc Fellowship.
\appendix
\section*{Appendix}
\begin{appendices}
\section{Input Correspondence Structure}
Each input SIFT correspondence is represented by a 7-dimensional vector comprising of various elements. 
The first four dimensions correspond to the coordinates of the corresponding points in the two images, specifically ($x_1$, $y_1$) and ($x_2$, $y_2$).
An additional dimension is derived from the Second Nearest Neighbor (SNN) ratio, which can be interpreted as an indicator of the matching quality.
Furthermore, we incorporate scale ($q \in \mathbb{R}$) and rotation ($\alpha \in [0, 2\pi]$) values that are derived from the image features. 
Specifically, the scale value, $q$, represents the ratio of the feature sizes in the two images and is calculated as $q = q_2 / q_1$. 
Here, $q_i$ denotes the feature size in the $i$th image.
Similarly, the rotation value, $\alpha$, represents the relative rotation from the first to the second image and is calculated as $\alpha = \alpha_2 - \alpha_1$, where $\alpha_i$ denotes the orientation in the $i$th image.
Hence, these parameters can be combined to form a 7-dimensional vector represented as $[x_1, y_1, x_2, y_2, \text{SNN}, q, \alpha]$. 

\end{appendices}
{\small
\bibliographystyle{ieee_fullname}
\bibliography{egbib}

\begin{thebibliography}{10}\itemsep=-1pt

\bibitem{tutorialdata}
Ransac tutorial 2020 data.
\newblock \url{https://github.com/ducha-aiki/ransac-tutorial-2020-data}.

\bibitem{arandjelovic2012three}
Relja Arandjelovi{\'c} and Andrew Zisserman.
\newblock Three things everyone should know to improve object retrieval.
\newblock In {\em CVPR}, 2012.

\bibitem{barath2018five}
Daniel Barath.
\newblock Five-point fundamental matrix estimation for uncalibrated cameras.
\newblock In {\em CVPR}, 2018.

\bibitem{barath2022learning}
Daniel Barath, Luca Cavalli, and Marc Pollefeys.
\newblock Learning to find good models in ransac.
\newblock In {\em CVPR}, 2022.

\bibitem{cvpr2020ransactutorial}
D. Barath, T-J. Chin, Chum Ond{\v{r}}ej., D. Mishkin, R. Ranftl, and J. Matas.
\newblock {RANSAC} in 2020 tutorial.
\newblock In {\em CVPR}, 2020.

\bibitem{barath2018efficient}
Daniel Barath and Levente Hajder.
\newblock Efficient recovery of essential matrix from two affine
  correspondences.
\newblock {\em IEEE Transactions on Image Processing}, 2018.

\bibitem{barath2019homography}
Daniel Barath and Zuzana Kukelova.
\newblock Homography from two orientation-and scale-covariant features.
\newblock In {\em ICCV}, 2019.

\bibitem{barath2018graph}
Daniel Barath and Ji{\v{r}}{\'\i} Matas.
\newblock Graph-cut {RANSAC}.
\newblock In {\em CVPR}, 2018.

\bibitem{barath2018multi}
Daniel Barath and Jir{\i} Matas.
\newblock Multi-class model fitting by energy minimization and mode-seeking.
\newblock In {\em ECCV}, 2018.

\bibitem{barath2020magsac++}
Daniel Barath, Jana Noskova, Maksym Ivashechkin, and Jir{\i} Matas.
\newblock Magsac++, a fast, reliable and accurate robust estimator.
\newblock In {\em CVPR}, 2020.

\bibitem{barath2019magsac}
Daniel Barath, Jana Noskova, and Ji{\v{r}}{\'\i} Matas.
\newblock {MAGSAC}: marginalizing sample consensus.
\newblock In {\em CVPR}, 2019.

\bibitem{barath2021marginalizing}
Daniel Barath, Jana Noskova, and Jir{\i} Matas.
\newblock Marginalizing sample consensus.
\newblock {\em TPAMI}, 2021.

\bibitem{bay2006surf}
Herbert Bay, Tinne Tuytelaars, and Luc Van~Gool.
\newblock {SURF}: Speeded up robust features.
\newblock In {\em ECCV}, 2006.

\bibitem{bian2019evaluation}
Jia-Wang Bian, Yu-Huan Wu, Ji Zhao, Yun Liu, Le Zhang, Ming-Ming Cheng, and Ian
  Reid.
\newblock An evaluation of feature matchers for fundamental matrix estimation.
\newblock In {\em BMVC}, 2019.

\bibitem{brachmann2019neural}
Eric Brachmann and Carsten Rother.
\newblock Neural-guided ransac: Learning where to sample model hypotheses.
\newblock In {\em ICCV}, 2019.

\bibitem{chum2005matching}
Ond{\v{r}}ej Chum and Ji{\v{r}}{\'\i} Matas.
\newblock Matching with {PROSAC}-progressive sample consensus.
\newblock In {\em CVPR}, 2005.

\bibitem{detone2018superpoint}
Daniel DeTone, Tomasz Malisiewicz, and Andrew Rabinovich.
\newblock Superpoint: Self-supervised interest point detection and description.
\newblock In {\em CVPR}, 2018.

\bibitem{fan2021efficient}
Aoxiang Fan, Jiayi Ma, Xingyu Jiang, and Haibin Ling.
\newblock Efficient deterministic search with robust loss functions for
  geometric model fitting.
\newblock {\em TPAMI}, 2021.

\bibitem{RANSAC}
Martin~A Fischler and Robert~C Bolles.
\newblock Random sample consensus: a paradigm for model fitting with
  applications to image analysis and automated cartography.
\newblock {\em Communications of the ACM}, 1981.

\bibitem{geiger2012we}
Andreas Geiger, Philip Lenz, and Raquel Urtasun.
\newblock Are we ready for autonomous driving? the {KITTI} vision benchmark
  suite.
\newblock In {\em CVPR}, 2012.

\bibitem{hartley2003multiple}
Richard Hartley and Andrew Zisserman.
\newblock {\em Multiple view geometry in computer vision}.
\newblock Cambridge university press, 2003.

\bibitem{he2015delving}
Kaiming He, Xiangyu Zhang, Shaoqing Ren, and Jian Sun.
\newblock Delving deep into rectifiers: Surpassing human-level performance on
  imagenet classification.
\newblock In {\em ICCV}, 2015.

\bibitem{Huang_2017_CVPR}
Gao Huang, Zhuang Liu, Laurens van~der Maaten, and Kilian~Q. Weinberger.
\newblock Densely connected convolutional networks.
\newblock In {\em CVPR}, 2017.

\bibitem{isack2012energy}
Hossam Isack and Yuri Boykov.
\newblock Energy-based geometric multi-model fitting.
\newblock {\em IJCV}, 2012.

\bibitem{ivashechkin2021vsac}
Maksym Ivashechkin, Daniel Barath, and Ji{\v{r}}{\'\i} Matas.
\newblock {VSAC}: Efficient and accurate estimator for {H} and {F}.
\newblock In {\em ICCV}, 2021.

\bibitem{kingma2014adam}
Diederik~P Kingma and Jimmy Ba.
\newblock Adam: A method for stochastic optimization.
\newblock {\em arXiv preprint arXiv:1412.6980}, 2014.

\bibitem{kipf2016semi}
Thomas~N Kipf and Max Welling.
\newblock Semi-supervised classification with graph convolutional networks.
\newblock {\em arXiv preprint arXiv:1609.02907}, 2016.

\bibitem{knapitsch2017tanks}
Arno Knapitsch, Jaesik Park, Qian-Yi Zhou, and Vladlen Koltun.
\newblock {T}anks and {T}emples: Benchmarking large-scale scene reconstruction.
\newblock {\em ACM Transactions on Graphics}, 2017.

\bibitem{lee2022self}
Jongmin Lee, Byungjin Kim, and Minsu Cho.
\newblock Self-supervised equivariant learning for oriented keypoint detection.
\newblock In {\em CVPR}, 2022.

\bibitem{lowe1999object}
D.~G. Lowe.
\newblock Object recognition from local scale-invariant features.
\newblock In {\em ICCV}, 1999.

\bibitem{lowe2004distinctive}
David~G Lowe.
\newblock Distinctive image features from scale-invariant keypoints.
\newblock {\em IJCV}, 2004.

\bibitem{ma2021image}
Jiayi Ma, Xingyu Jiang, Aoxiang Fan, Junjun Jiang, and Junchi Yan.
\newblock Image matching from handcrafted to deep features: A survey.
\newblock {\em IJCV}, 2021.

\bibitem{matas2004robust}
J. Matas, Chum Ond{\v{r}}ej., M. Urban, and T. Pajdla.
\newblock Robust wide-baseline stereo from maximally stable extremal regions.
\newblock {\em IVC}, 2004.

\bibitem{mikolajczyk2005comparison}
Krystian Mikolajczyk, Tinne Tuytelaars, Cordelia Schmid, Andrew Zisserman,
  Jir{\i} Matas, Frederik Schaffalitzky, Timor Kadir, and Luc Van~Gool.
\newblock A comparison of affine region detectors.
\newblock {\em IJCV}, 2005.

\bibitem{mishkin2015mods}
Dmytro Mishkin, Jir{\i} Matas, and Michal Perdoch.
\newblock {MODS}: Fast and robust method for two-view matching.
\newblock {\em Computer Vision and Image Understanding}, 2015.

\bibitem{mur2015orb}
Raul Mur-Artal, Jose Maria~Martinez Montiel, and Juan~D Tardos.
\newblock {ORB-SLAM}: a versatile and accurate monocular slam system.
\newblock {\em IEEE transactions on robotics}, 2015.

\bibitem{ni2009groupsac}
Kai Ni, Hailin Jin, and Frank Dellaert.
\newblock {GroupSAC}: Efficient consensus in the presence of groupings.
\newblock In {\em ICCV}. IEEE, 2009.

\bibitem{pham2014interacting}
Trung~Thanh Pham, Tat-Jun Chin, Konrad Schindler, and David Suter.
\newblock Interacting geometric priors for robust multimodel fitting.
\newblock {\em IEEE Transactions on Image Processing}, 2014.

\bibitem{pritchett1998wide}
P. Pritchett and A. Zisserman.
\newblock Wide baseline stereo matching.
\newblock In {\em ICCV}. IEEE, 1998.

\bibitem{raguram2013usac}
Rahul Raguram, Ond{\v{r}}ej Chum, Marc Pollefeys, Jir{\i} Matas, and
  Jan-Michael Frahm.
\newblock {USAC}: a universal framework for random sample consensus.
\newblock {\em TPAMI}, 2013.

\bibitem{dfe2018}
Rene Ranftl and Vladlen Koltun.
\newblock Deep fundamental matrix estimation.
\newblock In {\em ECCV}, 2018.

\bibitem{raposo2016theory}
Carolina Raposo and Joao~P Barreto.
\newblock Theory and practice of structure-from-motion using affine
  correspondences.
\newblock In {\em CVPR}, 2016.

\bibitem{rousseeuw1984least}
Peter~J Rousseeuw.
\newblock Least median of squares regression.
\newblock {\em Journal of the American statistical association}, 1984.

\bibitem{rousseeuw2005robust}
Peter~J Rousseeuw and Annick~M Leroy.
\newblock {\em Robust regression and outlier detection}.
\newblock 2005.

\bibitem{rublee2011orb}
Ethan Rublee, Vincent Rabaud, Kurt Konolige, and Gary Bradski.
\newblock {ORB}: An efficient alternative to {SIFT} or {SURF}.
\newblock In {\em ICCV}, 2011.

\bibitem{schoenberger2016sfm}
Johannes~Lutz Sch\"{o}nberger and Jan-Michael Frahm.
\newblock Structure-from-motion revisited.
\newblock In {\em CVPR}, 2016.

\bibitem{schoenberger2016mvs}
Johannes~Lutz Sch\"{o}nberger, Enliang Zheng, Marc Pollefeys, and Jan-Michael
  Frahm.
\newblock Pixelwise view selection for unstructured multi-view stereo.
\newblock In {\em ECCV}, 2016.

\bibitem{snavely2006photo}
Noah Snavely, Steven~M Seitz, and Richard Szeliski.
\newblock Photo tourism: exploring photo collections in {3D}.
\newblock In {\em SIGGRAPH}. 2006.

\bibitem{sturm2012benchmark}
J{\"u}rgen Sturm, Nikolas Engelhard, Felix Endres, Wolfram Burgard, and Daniel
  Cremers.
\newblock A benchmark for the evaluation of {RGB-D SLAM} systems.
\newblock In {\em 2012 IEEE/RSJ International Conference on Intelligent Robots
  and Systems}. IEEE, 2012.

\bibitem{acne2020}
Weiwei Sun, Wei Jiang, Andrea Tagliasacchi, Eduard Trulls, and Kwang~Moo Yi.
\newblock Attentive context normalization for robust permutation-equivariant
  learning.
\newblock In {\em CVPR}, 2020.

\bibitem{thomee2016yfcc100m}
Bart Thomee, David~A Shamma, Gerald Friedland, Benjamin Elizalde, Karl Ni,
  Douglas Poland, Damian Borth, and Li-Jia Li.
\newblock Yfcc100m: The new data in multimedia research.
\newblock {\em Communications of the ACM}, 2016.

\bibitem{torr2002napsac}
Philip~Hilaire Torr, Slawomir~J Nasuto, and John~Mark Bishop.
\newblock {NAPSAC}: High noise, high dimensional robust estimation-it's in the
  bag.
\newblock In {\em BMVC}, 2002.

\bibitem{torr2002bayesian}
P.~H.~S. Torr.
\newblock Bayesian model estimation and selection for epipolar geometry and
  generic manifold fitting.
\newblock {\em IJCV}, 2002.

\bibitem{torr2000mlesac}
P.~H.~S. Torr and A. Zisserman.
\newblock {MLESAC}: A new robust estimator with application to estimating image
  geometry.
\newblock {\em Computer Vision and Image Understanding (CVIU)}, 2000.

\bibitem{ulyanov2016instance}
Dmitry Ulyanov, Andrea Vedaldi, and Victor Lempitsky.
\newblock Instance normalization: The missing ingredient for fast stylization.
\newblock {\em arXiv preprint arXiv:1607.08022}, 2016.

\bibitem{van2014renyi}
Tim Van~Erven and Peter Harremos.
\newblock R{\'e}nyi divergence and kullback-leibler divergence.
\newblock {\em IEEE Transactions on Information Theory}, 2014.

\bibitem{ventura2022p1ac}
Jonathan Ventura, Zuzana Kukelova, Torsten Sattler, and Daniel Barath.
\newblock {P1AC}: Revisiting absolute pose from a single affine correspondence.
\newblock In {\em ICCV}, 2023.

\bibitem{wilson2014robust}
Kyle Wilson and Noah Snavely.
\newblock Robust global translations with {1DSfM}.
\newblock In {\em ECCV}, 2014.

\bibitem{woo2021global}
Sanghyun Woo, Dahun KIm, Joon-Young Lee, and In~So Kweon.
\newblock Global context and geometric priors for effective non-local
  self-attention.
\newblock In {\em BMVC}, 2021.

\bibitem{cne2018}
Kwang~Moo Yi*, Eduard Trulls*, Yuki Ono, Vincent Lepetit, Mathieu Salzmann, and
  Pascal Fua.
\newblock Learning to find good correspondences.
\newblock In {\em CVPR}, 2018.

\bibitem{oanet2019}
Jiahui Zhang, Dawei Sun, Zixin Luo, Anbang Yao, Lei Zhou, Tianwei Shen, Yurong
  Chen, Long Quan, and Hongen Liao.
\newblock Learning two-view correspondences and geometry using order-aware
  network.
\newblock {\em ICCV}, 2019.

\bibitem{clnet2021}
Chen Zhao, Yixiao Ge, Feng Zhu, Rui Zhao, Hongsheng Li, and Mathieu Salzmann.
\newblock Progressive correspondence pruning by consensus learning.
\newblock In {\em ICCV}, 2021.

\end{thebibliography}
}

\end{document}